\documentclass{article}

\usepackage{geometry}
\usepackage{amsmath}
\usepackage{amsthm}
\usepackage{float}
\usepackage{amssymb}
\usepackage{graphicx}
\usepackage{hyperref}
\usepackage{xcolor}
\usepackage{esvect}
\usepackage[shortlabels]{enumitem}
\usepackage{ctable}
\usepackage{tabularx}
\usepackage{multirow}
\usepackage{booktabs}
\usepackage[caption=false]{subfig}
\usepackage{mlmath}

\usepackage{changepage}

\usepackage{authblk}

\makeatletter
    \def\blfootnote{\xdef\@thefnmark{}\@footnotetext}
\makeatother

\title{Deep frequency principle towards understanding why deeper learning is faster}
\author[a*]{Zhi-Qin John Xu}
\author[a]{Hanxu Zhou}
\affil[a]{School of Mathematical Sciences, MOE-LSC and Institute of Natural Sciences, Shanghai Jiao Tong University, Shanghai, 200240, P.R. China}
\blfootnote{* xuzhiqin@sjtu.edu.cn. To Appear in AAAI-2021.}
\date{\today}

\begin{document}
\maketitle

\begin{abstract}
Understanding the effect of depth in deep learning is a critical problem. In this work, we utilize the Fourier analysis to empirically provide a promising mechanism to understand why feedforward deeper learning is faster. To this end,  we separate a deep neural network, trained by normal stochastic gradient descent, into two parts during analysis, i.e., a  pre-condition component and  a learning component, in which the output of the pre-condition one is the input of the learning one. We use a filtering method to characterize the frequency distribution of a high-dimensional function.  Based on experiments of deep networks and real dataset, we propose a deep frequency principle, that is, the effective target function for a deeper hidden layer biases towards  lower  frequency during the training. Therefore, the learning component effectively learns a lower frequency function if the pre-condition component has more layers. Due to the well-studied frequency principle, i.e., deep neural networks learn lower frequency functions faster, the deep frequency principle provides a reasonable explanation to why deeper learning is faster. We believe these empirical studies would be valuable for future theoretical studies of the effect of depth in deep learning.
\end{abstract}

\section{Introduction}

\begin{figure} 
    \begin{centering}
    \subfloat[different networks]{
        \includegraphics[scale=0.52]{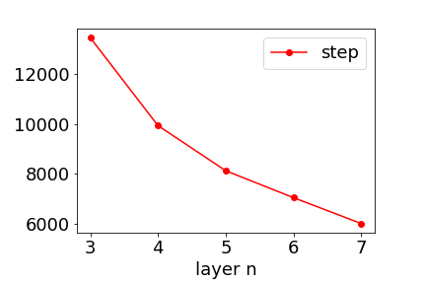}
    }
    \subfloat[different target functions]{
        \includegraphics[scale=0.44]{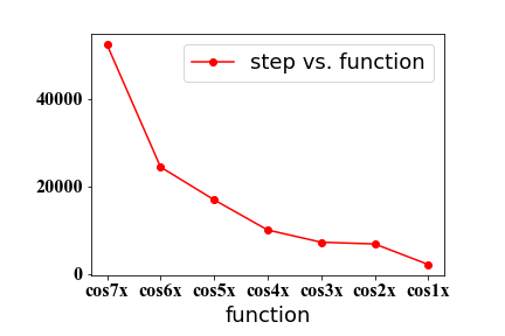}
    }
    \par\end{centering}
    \caption{Training epochs (indicated by ordinate axis) of different deep neural networks when they achieve a fixed error. (a) Using networks with different number of hidden layers with the same size to learn data sampled from a target function $\cos(3x)+\cos(5x)$. (b) Using a fixed network to learn data sampled from different target functions.}\label{fig:1d}
\end{figure}

Deep neural networks  have achieved tremendous success in many
applications, such as computer vision, speech recognition, speech
translation, and natural language processing etc. The depth in neural networks plays an important role in the applications. Understanding the effect of depth is a central problem to reveal the ``black box'' of deep learning. For example, empirical studies show that a deeper network can learn faster and generalize better in both real data and synthetic data \cite{he2016deep,arora2018optimization}. Different network structures have different computation costs in each training epoch. In this work, we define that \emph{the learning of a deep neural network is faster if the loss of the deep neural network decreases to a designated error with fewer training epochs.}    For example, as shown in Fig.~\ref{fig:1d} (a), when learning data sampled from a target function $\cos(3x)+\cos(5x)$, a deep neural network with more hidden layers achieves the designated training loss with fewer training epochs. Although empirical studies suggest deeper neural networks may learn faster, there is few understanding of the mechanism.

In this work, we would empirically explore an underlying mechanism that may explain why deeper neural network (note: in this work, we only study feedforward networks) can learn faster from the perspective of Fourier analysis. We start from a universal phenomenon of frequency principle \cite{xu_training_2018,rahaman2018spectral,xu2019frequency,luo2019theory,e2019machine}, that is, deep neural networks often fit target
functions from low to high frequencies during the training. Recent works show that frequency principle may provide an understanding  to the success and failure of deep learning \cite{xu2019frequency,zhang2019explicitizing,e2019machine,ma2020slow}. We use an ideal example to illustrate the frequency principle, i.e., using a deep neural network to fit different target functions. As the frequency of the target function decreases, the deep neural network achieves a designated error with fewer training epochs, which is similar to the phenomenon when using a deeper network to learn a fixed target function. 

\begin{figure}[ht]
    \begin{centering}
    \includegraphics[scale=0.36]{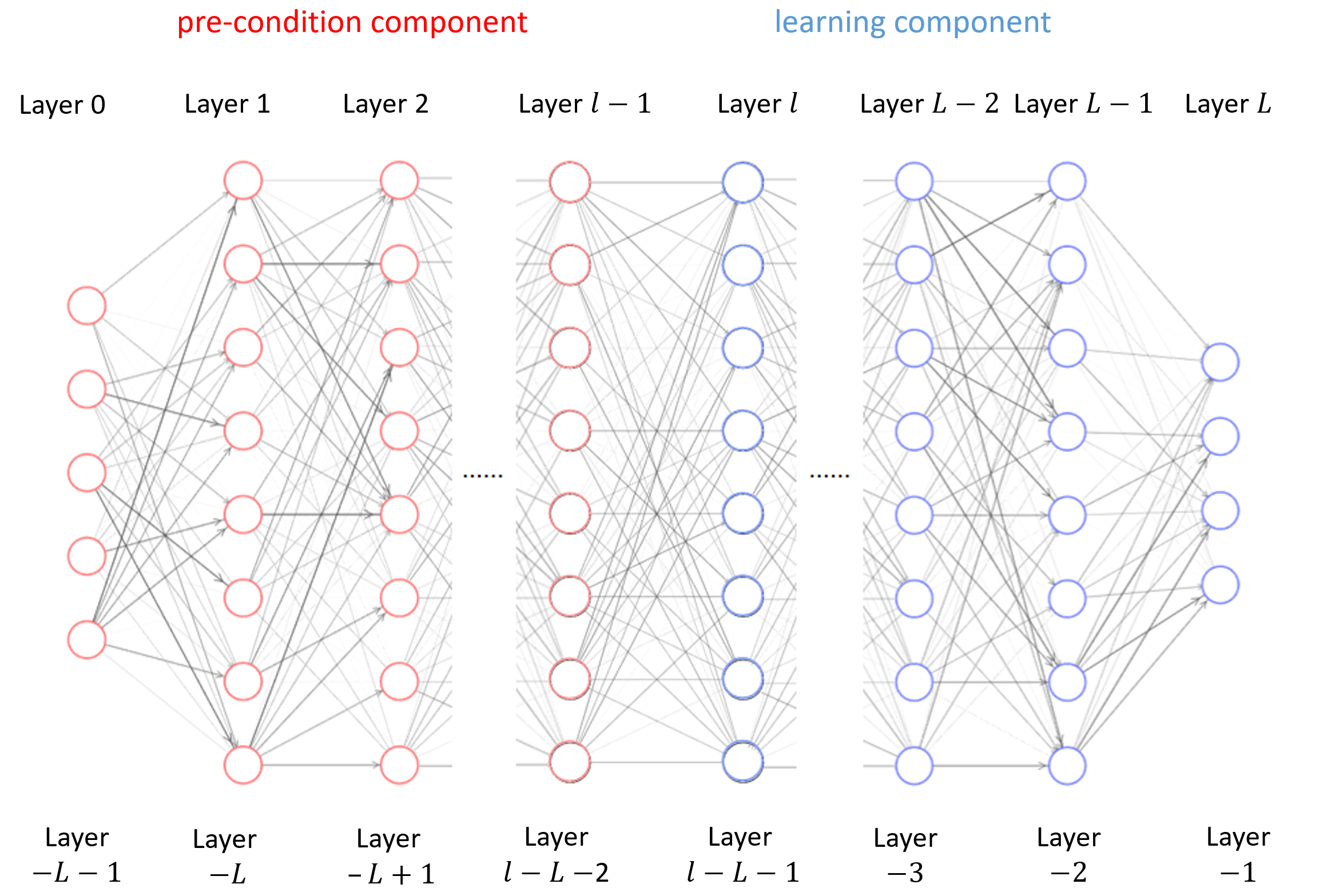}
    \par\end{centering}
    \caption{General deep neural network.}\label{fig:net}
\end{figure}

Inspired by the above analysis, we propose a mechanism to understand why a deeper network, $f_\theta(\vx)$, faster learns a set of training data,  $S=\{(\vx_i,y_i)\}_{i=1}^{n}$ sampled from a target function $f^{*}(\vx)$, illustrated as follows. Networks are trained as usual while we separate a deep neural network into two parts in the analysis, as shown in Fig. \ref{fig:net}, one is a pre-condition component and the other is a learning component, in which the output of the pre-condition one, denoted as $f_{\vtheta}^{[l-1]}(\vx)$ (first $l-1$ layers are classified as the pre-condition component), is the input of the learning one. For the learning component, the effective training data at each training epoch is $S^{[l-1]}=\{(f_{\vtheta}^{[l-1]}(\vx_i),y_i)\}_{i=1}^{n}$. We then perform experiments based on the variants of Resnet18  structure \cite{he2016deep} and CIFAR10 dataset. We fix the learning component (fully-connected layers). When increasing the number of the pre-condition layer (convolution layers), we find that $S^{[l-1]}$ has a stronger bias towards low frequency during the training. By frequency principle, the learning of a lower frequency function is faster, therefore, the learning component is faster to learn $S^{[l-1]}$ when the pre-condition component has more layers. The analysis among different network structures is often much more difficult than the analysis of one single structure. For providing hints for future theoretical study, we study a fixed fully-connected deep neural network by classifying different number of layers into the pre-condition component, i.e., varying $l$ for a network in the analysis. As $l$ increases, we similarly find that $S^{[l]}$ contains more low frequency and less high frequency during the training. Therefore, we propose the following principle: 

\begin{changemargin}{0.5cm}{0.5cm}
    \emph{Deep frequency principle: The effective target function for a deeper hidden layer biases towards  lower  frequency during the training.}
\end{changemargin}

With the well-studied frequency principle, the deep frequency principle shows a promising mechanism  for understanding why a deeper network learns faster.

\subsection{Related work}
From the perspective of approximation, the expressive power of a deep neural network increases with the depth \cite{telgarsky2016benefits,eldan2016power,weinan2018exponential}. However, the approximation theory renders no implication on the optimization of deep neural networks. 

With residual connection, \cite{he2016deep} successfully train very deep networks and find that deeper networks can achieve better generalization error. In addition,  \cite{he2016deep} also show that the training of deeper network is faster. \cite{arora2018optimization} show that the acceleration effect of depth also exists in deep linear neural network and provide a viewpoint for understanding the effect of depth, that is, increasing depth can be seen as an acceleration procedure that combines momentum with adaptive learning rates. There are also many works studying the effect of depth for deep linear networks \cite{saxe2014exact,kawaguchi2019effect,gissin2019implicit,shin2019effects}. In this work, we study the optimization effect of depth in non-linear deep networks.

Various studies suggest that the function learned by the deep neural networks increases its complexity as the training goes \cite{arpit2017closer,valle2018deep,mingard2019neural,kalimeris2019sgd,yang2019fine}.
This increasing complexity is also found in deep linear network \cite{gissin2019implicit}. The high-dimensional experiments in \cite{xu2019frequency} show that the low-frequency part is converged first, i.e., frequency principle. Therefore, the ratio of the power of the low-frequency component of the deep neural network output experiences a increasing stage at the beginning (due to the convergence of low-frequency part), followed by a decreasing stage (due to the convergence of high-frequency part). As more high-frequency involved, the complexity of the deep neural network output increases. Therefore, the ratio of the low-frequency component used in this paper validates the complexity increasing during the training, which is consistent with other studies.

Frequency principle is examined in extensive datasets and deep neural networks \cite{xu_training_2018,rahaman2018spectral,xu2019frequency}. Theoretical studies subsequently shows that frequency principle holds in general setting with infinite samples \cite{luo2019theory} and in the regime of wide neural networks (Neural Tangent Kernel (NTK) regime \cite{jacot2018neural}) with finite samples \cite{zhang2019explicitizing} or sufficient many samples \cite{cao2019towards,yang2019fine,basri2019convergence,bordelon2020spectrum}. \cite{e2019machine} show that the integral equation would naturally leads to the frequency principle. With the theoretical understanding, the frequency principle inspires the design of deep neural networks to fast learn a function with high frequency \cite{liu2020multi,wang2020multi,jagtap2019adaptive,cai2019phasednn,biland2019frequency,li2020multi}. 

\section{Preliminary}
\subsection{Low Frequency Ratio (LFR)} 
To compare two 1-d functions in the frequency domain, we can display their spectrum. However, this does not apply for high-dimensional functions, because the computation cost of high-dimensional Fourier transform suffers from the curse of dimensionality. To overcome this, we use a low-frequency filter to derive a low-frequency component of the interested function and then use a Low Frequency Ratio (LFR) to characterize the power ratio of the low-frequency component over the whole spectrum. 

The LFR is defined as follows. We first split the frequency domain into two parts, i.e., a low-frequency
part with frequency $|\vec{k}|\leq k_{0}$ and a high-frequency part with
$|\vec{k}|>k_{0}$, where $|\cdot|$ is the length of a vector.
Consider a dataset $\{(\vec{x}_{i},\vec{y}_{i})\}_{i=1}^{n}$, $\vx_i\in R^{d}$, and $\vy_i\in R^{d_o}$. For example, $d=784$ and $d_o=10$ for MNIST and $d=3072$ and $d_o=10$ for CIFAR10. The LFR is defined as 
\begin{equation}
    \mathrm{LFR}(k_0)=\frac{\sum_{\vec{k}}\mathbbm{1}_{|\vec{k}|\leq k_{0}}|\vec{\hat{y}}(\vec{k})|^{2}}{\sum_{\vec{k}}|\vec{\hat{y}}(\vec{k})|^{2}},
\end{equation}
 where $\hat{\cdot}$ indicates Fourier transform, $\mathbbm{1}_{\vec{k}\leq k_{0}}$
is an indicator function, i.e., 
\begin{equation*}
    \mathbbm{1}_{|\vec{k}|\leq k_{0}}=\begin{cases}
    1, & |\vec{k}|\leq k_{0},\\
    0, & |\vec{k}|>k_{0}.
    \end{cases}
\end{equation*}

However, it is almost impossible to compute above quantities
numerically due to high computational cost of high-dimensional Fourier
transform. Similarly as previous study \cite{xu2019frequency}, We alternatively use the Fourier transform of a Gaussian
function $\hat{G}^{\delta}(\vec{k})$, where $\delta$ is the variance
of the Gaussian function $G$, to approximate $\mathbbm{1}_{|\vec{k}|>k_{0}}$. Note that $1/\delta$ can be interpreted as the variance of $\hat{G}$. 
The approximation is reasonable due to the following two reasons. First, the Fourier
transform of a Gaussian is still a Gaussian, i.e., $\hat{G}^{\delta}(\vec{k})$
decays exponentially as $|\vec{k}|$ increases, therefore, it can
approximate $\mathbbm{1}_{|\vec{k}|\leq k_{0}}$ by $\hat{G}^{\delta}(\vec{k})$
with a proper $\delta(k_{0})$.
Second, the computation of $\mathrm{LFR}$
contains the multiplication of Fourier transforms in the frequency
domain, which is equivalent to the Fourier transform of a convolution
in the spatial domain. We can equivalently perform the computation
in the spatial domain so as to avoid the almost impossible high-dimensional
Fourier transform. The low frequency part can be derived by 
\begin{equation}
    \vec{y}_{i}^{\mathrm{low},\delta(k_0)}\triangleq(\vec{y}*G^{\delta(k_0)})_{i},\label{eq:filter-1}
\end{equation}
where $*$ indicates convolution operator.
Then, we can compute the LFR by 
\begin{equation}
     \mathrm{LFR}(k_0)=\frac{\sum_{i}|\vec{y}_{i}^{\mathrm{low},\delta(k_0)}|^{2}}{\sum_{i}|\vec{y}_{i}|^{2}}.
\end{equation}
The low frequency part can be derived on the discrete data points
by 
\begin{equation}
    \vec{y}_{i}^{{\rm low},\delta}=\frac{1}{C_{i}}\sum_{j=0}^{n-1}\vec{y}_{j}G^{\delta}(\vec{x}_{i}-\vec{x}_{j}),\label{eq:filter}
\end{equation}
where $C_{i}=\sum_{j=0}^{n-1}G^{\delta}(\vec{x}_{i}-\vec{x}_{j})$
is a normalization factor and 
\begin{equation}
    G^{\delta}(\vec{x}_{i}-\vec{x}_{j})=\exp\left(-|\vec{x}_{i}-\vec{x}_{j}|^{2}/2\delta\right).
\end{equation}
$1/\delta$ is the variance of $\hat{G}$, therefore, it can be interpreted as the frequency width outside which is filtered out by convolution.

\subsection{Ratio Density Function (RDF) }
\begin{figure}[ht]
    \begin{centering}
    \subfloat[]{
        \includegraphics[scale=0.5]{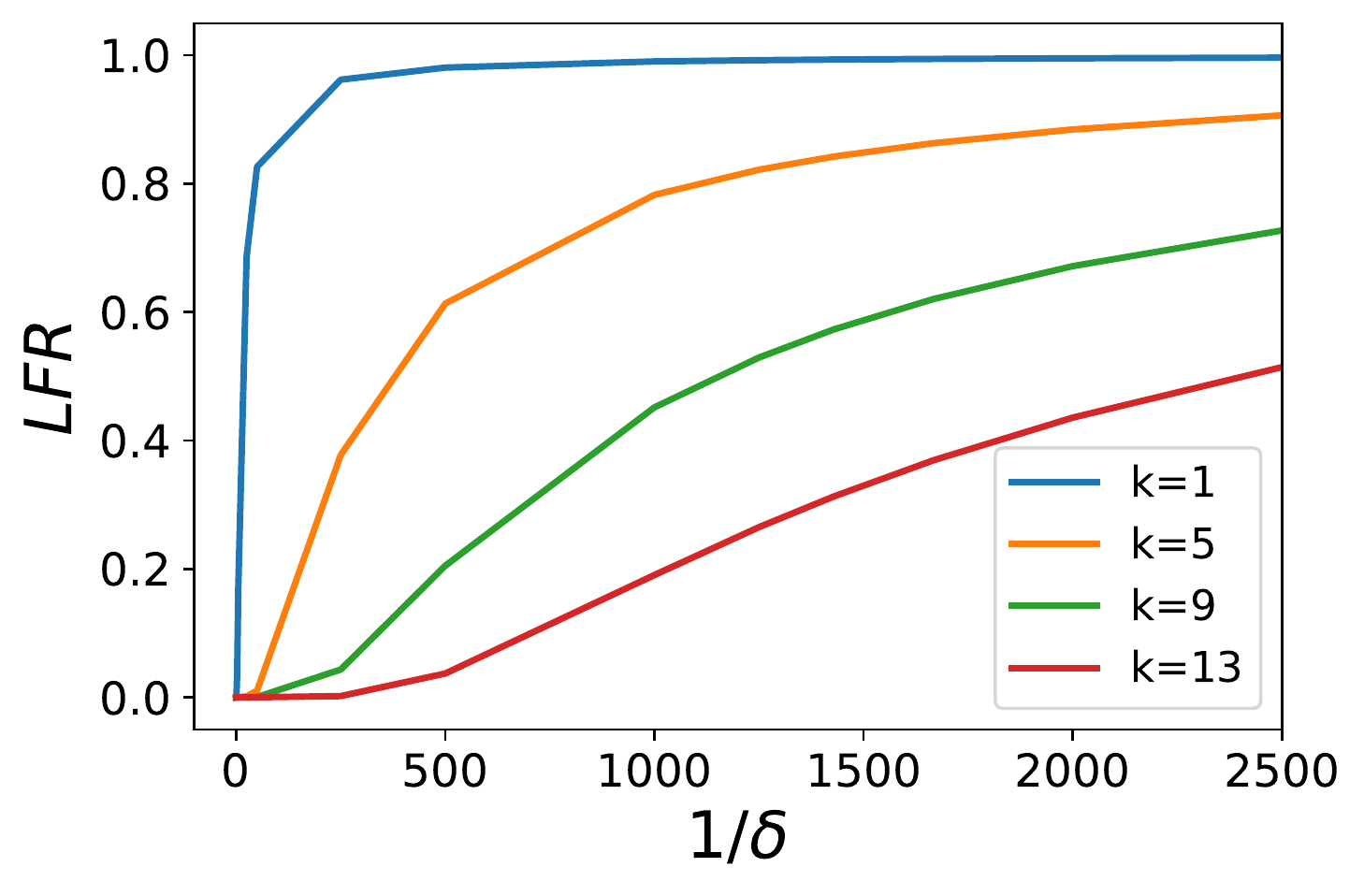}
    } \subfloat[]{
        \includegraphics[scale=0.5]{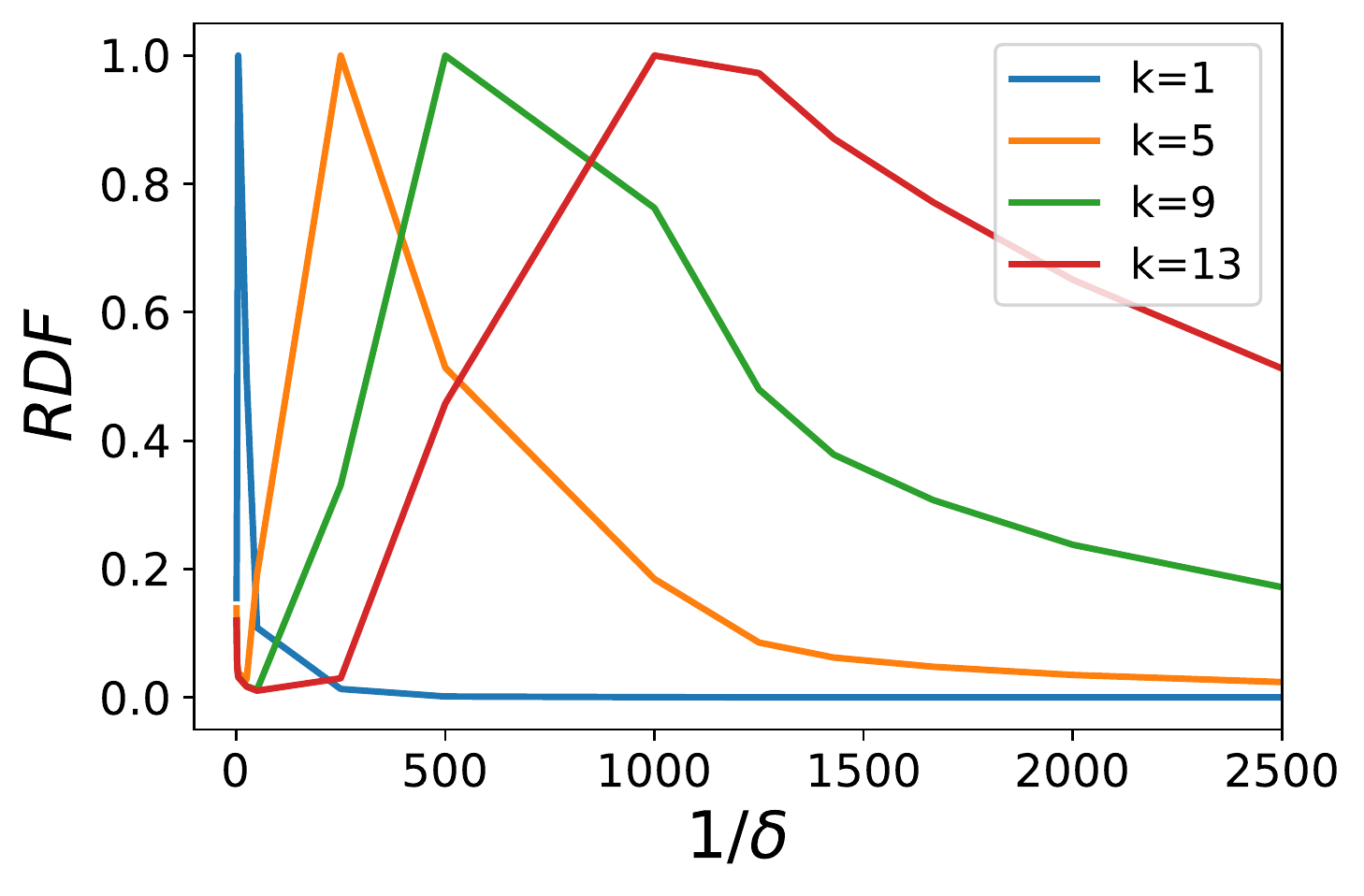}
    }
    \par\end{centering}
    \caption{LFR and RDF for $\sin (k\pi x)$ vs. $1/\delta$. Note that we normalize RDF in (b) by the maximal value of each curve for visualization.}\label{fig:sin}
\end{figure}

$\mathrm{LFR}(k_0)$ characterizes the power ratio of frequencies within a sphere of radius $k_0$. To characterize each frequency in the radius direction, similarly to probability, we define the ratio density function (RDF) as 
\begin{equation}
    {\rm RDF}(k_0)= \frac{\partial  \mathrm{LFR}(k_0)}{\partial k_0}.
\end{equation}
In practical computation, we use $1/\delta$ for $k_0$ and use the linear slope between two consecutive points for the derivative. For illustration, we show the LFR and RDF for $\sin(k\pi x)$ in Fig. \ref{fig:sin}. As shown in Fig. \ref{fig:sin}(a), the LFR of low-frequency function faster approaches one when the filter width in the frequency domain is small, i.e., small $1/\delta$. The RDF in Fig. \ref{fig:sin}(b)  shows that as $k$ in the target function increases, the peak of RDF moves towards wider filter width, i.e., higher frequency. Therefore, it is more intuitive that the RDF effectively reflects where the power of the function concentrates in the frequency domain. In the following, we will use RDF to study the frequency distribution of effective target functions for hidden layers.
\subsection{General deep neural network}
We adopt the suggested standard notation in \cite{beijing2020Suggested}. An $L$-layer neural network is defined recursively,
\begin{align}
     & f_{\vtheta}^{[0]}(\vx)=\vx,                                                                                   \\
     & f_{\vtheta}^{[l]}(\vx)=\sigma\circ(\mW^{[l-1]} f_{\vtheta}^{[l-1]}(\vx) + \vb^{[l-1]}) \quad 1\leq l\leq L-1, \\
     & f_{\vtheta}(\vx)=f_{\vtheta}^{[L]}(\vx)=\mW^{[L-1]} f_{\vtheta}^{[L-1]}(\vx) + \vb^{[L-1]},
\end{align}
where $\mW^{[l]} \in \sR^{m_{l+1}\times m_{l}}$, $\vb^{[l]}=\sR^{m_{l+1}}$, $m_0=d_{\rm in}=d$, $m_{L}=d_{\rm o}$,
$\sigma$ is a scalar function and ``$\circ$'' means entry-wise operation.
We denote the set of parameters by $\vtheta$.
For simplicity, we also denote 
\begin{equation}
    f_{\vtheta}^{[-l]}(\vx)=f_{\vtheta}^{[L-l+1]}(\vx). 
\end{equation}
For example, the output layer is layer ``$-1$'', i.e., $f_{\vtheta}^{[-1]}(\vx)$ for a given input $\vx$, and the last hidden layer is layer ``$-2$'', i.e., $f_{\vtheta}^{[-2]}(\vx)$ for a given input $\vx$, illustrated in Fig. \ref{fig:net}.

The effective target function for the learning component, consisting from layer ``$l$'' to the output layer, is 
\begin{equation}
    S^{[l-1]}=\{(f_{\vtheta}^{[l-1]}(\vx_i),y_i)\}_{i=1}^{n}.
\end{equation}

\subsection{Training details} \label{training}
We list training details for experiments as follows.

\begin{figure} 
    \begin{centering}
    \includegraphics[scale=0.27]{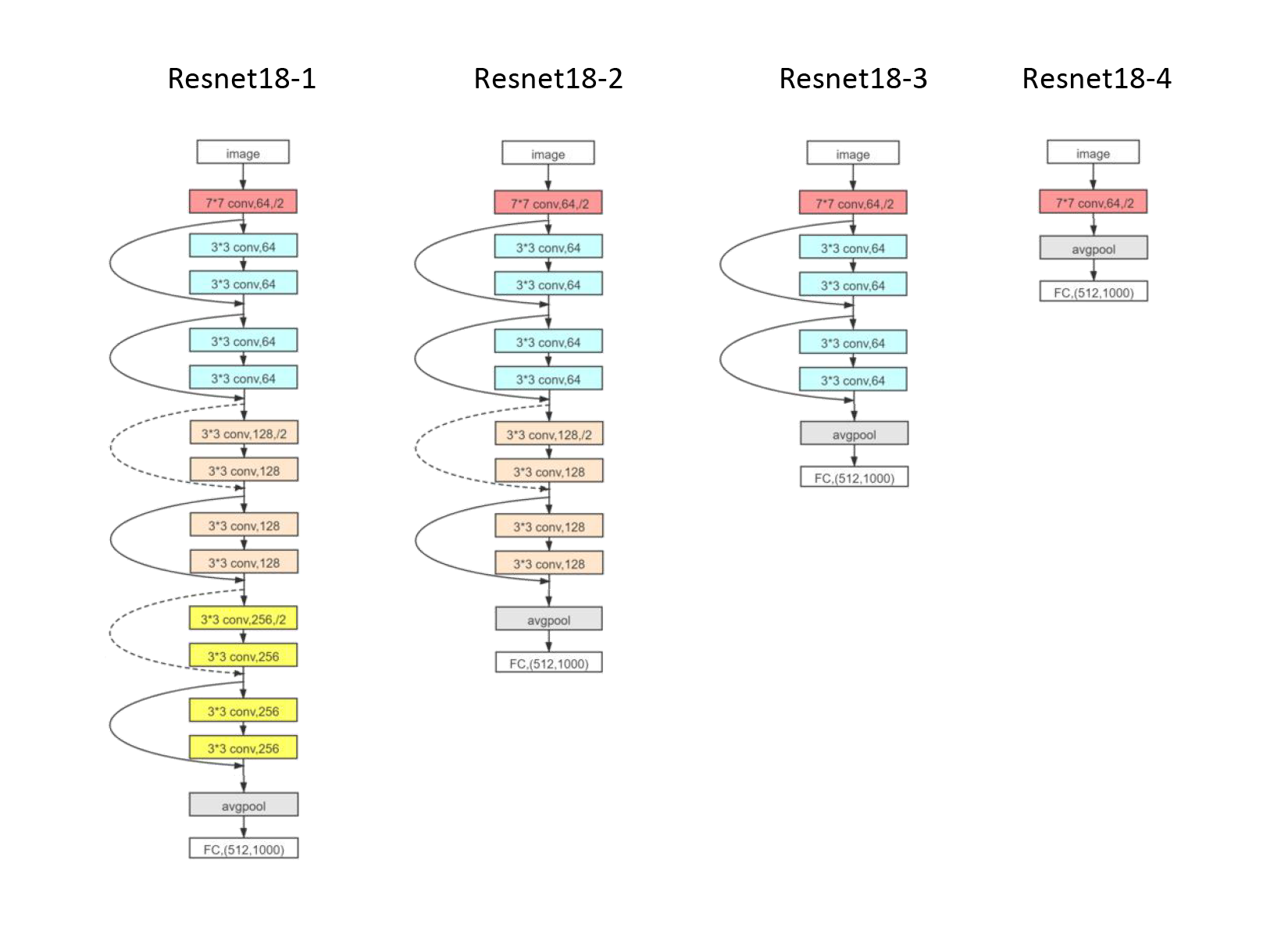}
    \par\end{centering}
    \caption{Variants of Resnet18.}\label{fig:resnet}
\end{figure}

\begin{figure}[ht]
    \begin{centering}
    \subfloat[Training]{
        \includegraphics[scale=0.56]{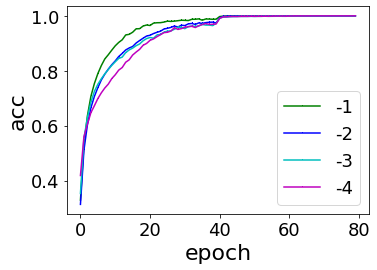}
    }
    \subfloat[Validation]{
        \includegraphics[scale=0.56]{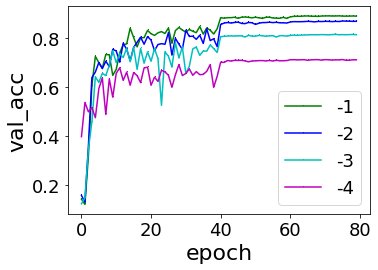}
    }
    \par\end{centering}
    \caption{Training accuracy and validation accuracy vs. epoch for variants of Resnet18.}\label{fig:resnetacc}
\end{figure}

\begin{figure*} 
    \begin{centering}
    \subfloat[epoch 0]{
        \includegraphics[scale=0.34]{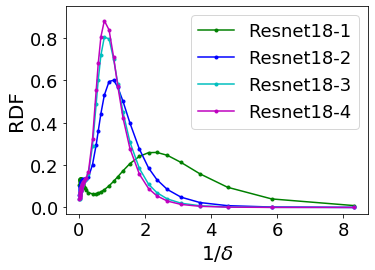}
    }
    \subfloat[epoch 1]{
        \includegraphics[scale=0.34]{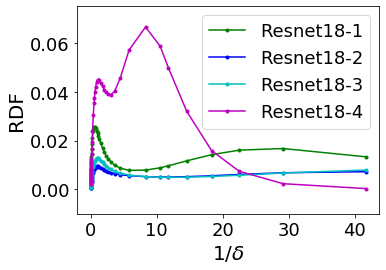}
    }\subfloat[epoch 2]{
        \includegraphics[scale=0.34]{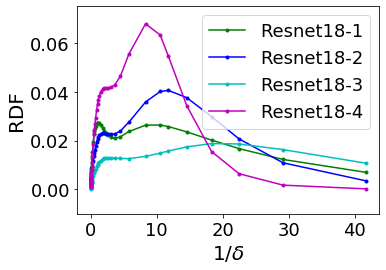}
    }
    
    \subfloat[epoch 3]{
        \includegraphics[scale=0.34]{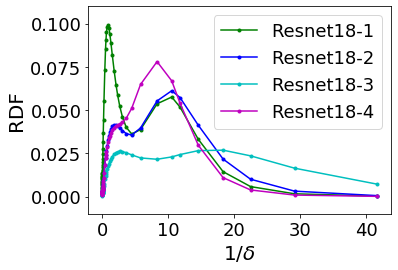}
    }
    \subfloat[epoch 15]{
        \includegraphics[scale=0.34]{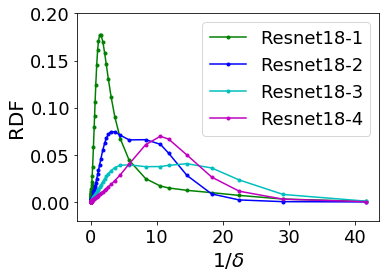}
    }
    \subfloat[epoch 80]{
        \includegraphics[scale=0.34]{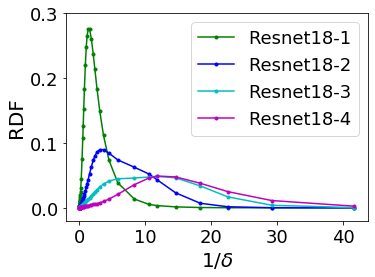}
    }
    \par\end{centering}
    \caption{RDF of $S^{[-3]}$ (effective target function of layer ``-2'') vs. $1/\delta$ at different epochs for variants of Resnet18.}\label{fig:resnetpdf}
\end{figure*}

For the experiments of the variants of Resnet18 on CIFAR10, the network structures are shown in Fig. \ref{fig:resnet}. The output layer is equipped with softmax and the network is trained by Adam optimizer with cross-entropy loss and batch size $256$. The learning rate is changed as the training proceeds, that is, $10^{-3}$ for epoch 1-40 , $10^{-4}$ for epoch 41-60, and $10^{-5}$ for epoch 61-80. We use 40000 samples of  CIFAR10 as the training set and 10000 examples as the validation set. The training accuracy and the validation accuracy are shown in Fig. \ref{fig:resnetacc}. The RDF of the effective target function of the last hidden layer for each variant is shown in Fig. \ref{fig:resnetpdf}.

For the experiment of fully-connected network on MNIST, we choose the activation function of $\tanh$ and size $784-500-500-500-500-500-10$. The output layer of the network does not equip any activation function. The network is trained by Adam optimizer with mean squared loss, batch size $256$ and learning rate $10^{-5}$. The training is stopped when the loss is smaller than $10^{-2}$. We use 30000 samples of the MNIST as training set. The RDF of the effective target functions of different hidden layers are shown in Fig. \ref{fig:dnnmnist}.

Note that ranges of different dimensions in the input are different, which would result in that for the same $\delta$, different dimensions keep different frequency ranges when convolving with the Gaussian function. Therefore, we normalized each dimension by its maximum amplitude, thus, each dimension lies in $[-1,1]$. Without doing such normalization, we still obtain similar results of deep frequency principle.

All codes are written by Python and Tensorflow, and  run on Linux system with Nvidia GTX 2080Ti or Tesla V100 cards. Codes can be found at \url{github.com}.

\section{Results}
Based on the experiments of deep networks and real datasets, we would show a deep frequency principle, a promising mechanism, to understand why deeper neural networks learn faster, that is, the effective target function for a deeper hidden layer biases towards  lower  frequency during the training. To derive the effective function, we decompose the target function into a pre-condition component, consisting of layers before the considered hidden layer, and a learning component, consisting from the considered hidden layer to the output layer, as shown in Fig. \ref{fig:net}. As the considered hidden layer gets deeper, the learning component effectively learns a lower frequency function. Due to the frequency principle, i.e., deep neural networks learn low frequency faster, a deeper neural network can learn the target function faster. The key to validate deep frequency principle is to show the frequency distribution of the effective target function for each considered hidden layer. 

First, we study a practical and common situation, that is, networks with more hidden layers learn faster. Then, we examine the deep frequency principle on a fixed deep neural network but consider the effective target function for different hidden layers. 

\begin{figure*}
    \begin{centering}
    \subfloat[epoch 0]{
        \includegraphics[scale=0.34]{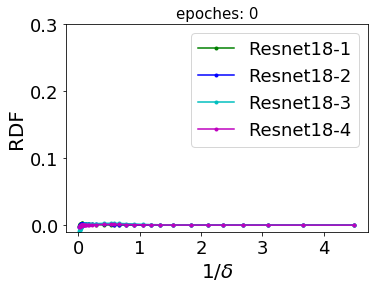}
    }
    \subfloat[epoch 1]{
        \includegraphics[scale=0.34]{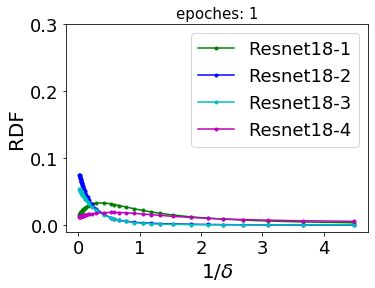}
    }\subfloat[epoch 2]{
        \includegraphics[scale=0.34]{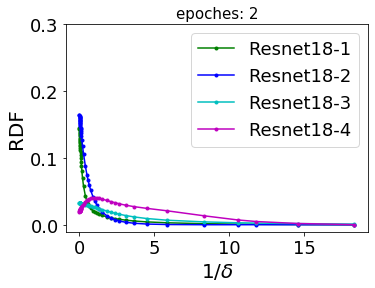}
    }
    
    \subfloat[epoch 3]{
        \includegraphics[scale=0.34]{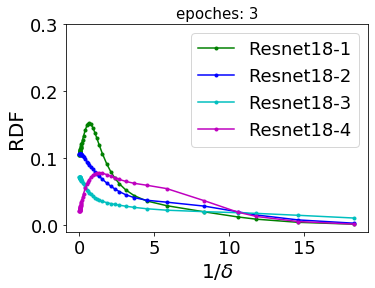}
    }
    \subfloat[epoch 15]{
        \includegraphics[scale=0.34]{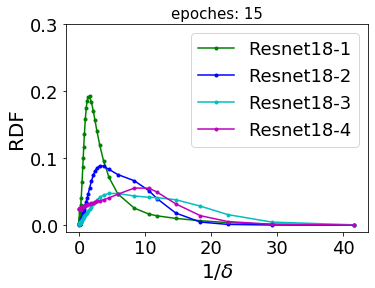}
    }
    \subfloat[epoch 80]{
        \includegraphics[scale=0.34]{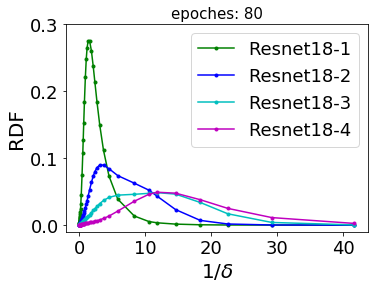}
    }
    \par\end{centering}
    \caption{RDF of  $\left\{\left(f_{\vtheta}^{[-3]}(\vx_i),f_{\vtheta}^{[-1]}(\vx_i)\right)\right\}_{i=1}^{n}$  vs. $1/\delta$ at different epochs for variants of Resnet18.}\label{fig:appendresnetpdf}
\end{figure*}

\subsection{Deep frequency principle on variants of Resnet18}
In this subsection, we would utilize variants of Resnet18 and CIFAR10 dataset to validate deep frequency principle. The structures of four variants are illustrated as follows. As shown in Fig. \ref{fig:resnet}, all structures have several convolution parts, followed by two same fully-connected layers. Compared with Resnet18-$i$, Resnet18-$(i+1)$ drops out a  convolution part and keep other parts the same. 

As shown in Fig. \ref{fig:resnetacc}, a deeper net attains a fixed training accuracy with fewer training epochs and achieves a better generalization after training.

From the layer ``-2'' to the final output, it can be regarded as a two-layer neural network, which is widely studied. Next, we examine the RDF for layer ``-2''. The effective target function is
\begin{equation}
    S^{[-3]}=\left\{\left(f_{\vtheta}^{[-3]}(\vx_i),\vy_i\right)\right\}_{i=1}^{n}.
\end{equation}
As shown in Fig. \ref{fig:resnetpdf}(a), at initialization, the RDFs for deeper networks concentrate at higher frequencies. However, as training proceeds, the concentration of RDFs of deeper networks moves towards lower frequency faster.  Therefore, for the two-layer neural network with a deeper pre-condition component, learning can be accelerated due to the fast convergence of low frequency in neural network dynamics, i.e., frequency principle.

For the two-layer neural network embedded as the learning component of the full network, the effective target function is $S^{[-3]}$. As the pre-condition component has more layers, layer ``-2'' is a  \emph{deeper} hidden layer in the full network. Therefore, Fig. \ref{fig:resnetpdf} validates that the effective target function for a deeper hidden layer biases towards  lower  frequency during the training, i.e., deep frequency principle. One may curious about how is the frequency distribution of the effective function of the learning component, i.e., $\{f_{\vtheta}^{[-3]}(\vx), f_{\vtheta}^{[-1]}(\vx)\}$. We consider RDF for the effective function evaluated on training points, that is, $\left\{\left(f_{\vtheta}^{[-3]}(\vx_i),f_{\vtheta}^{[-1]}(\vx_i)\right)\right\}_{i=1}^{n}$. This is similar as the effective target function, that is, those in deeper networks bias towards more low frequency function, as shown in Fig. \ref{fig:appendresnetpdf}. 

We have also performed many other experiments and validate the deep frequency principle, such as networks with different activation functions, fully-connected networks without residual connection, and different loss functions. An experiment for discussing the residual connection is presented in the discussion part.


The comparison in  experiments above crosses different networks, which would be difficult for future analysis. Alternatively, we can study how RDFs of $S^{[l]}$ of  different $l$ in a fixed deep network evolves during the training process. As expected, as $l$ increases, $S^{[l]}$ would be dominated by more lower frequency during the training process.

\subsection{RDF of different hidden layers in a fully-connected deep neural network}

\begin{figure*}[ht]
    \begin{centering}
    \subfloat[epoch 0]{
        \includegraphics[scale=0.28]{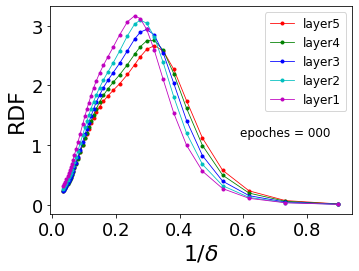}
    }
    \subfloat[epoch 100]{
        \includegraphics[scale=0.28]{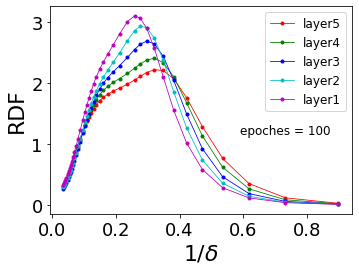}
    }
    \subfloat[epoch 200]{
        \includegraphics[scale=0.28]{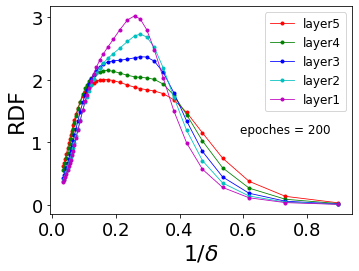}
    }
    \subfloat[epoch 300]{
        \includegraphics[scale=0.28]{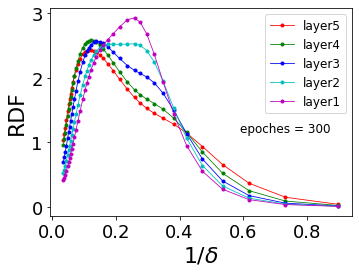}
    }
    
    \subfloat[epoch 400]{
        \includegraphics[scale=0.28]{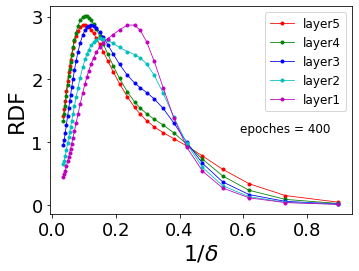}
    }
    \subfloat[epoch 600]{
        \includegraphics[scale=0.28]{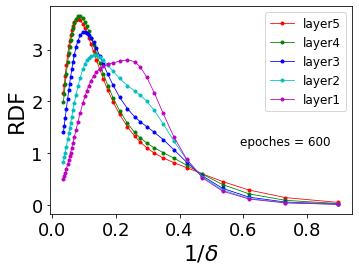}
    }
    \subfloat[epoch 800]{
        \includegraphics[scale=0.28]{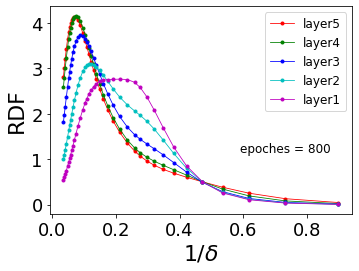}
    }
    \subfloat[epoch 900]{
        \includegraphics[scale=0.28]{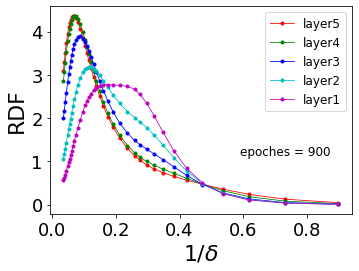}
    }
    \par\end{centering}
    \caption{RDF of different hidden layers vs. $1/\delta$ at different epochs for a fully-connected deep neural network when learning MNIST.  The five  colored curves are for five hidden layers, respectively. The curve with legend ``layer $i$'' is the RDF of $S^{[l]}$. }\label{fig:dnnmnist}
\end{figure*}

As analyzed above, an examination of the deep frequency principle in a deep neural network would provide valuable insights for future theoretical study. A key problem is that different hidden layers often have different sizes, i.e., $S^{[l]}$'s have different input dimensions over different $l$'s. LFR is similar to a volume ratio, thus, it depends on the dimension. To control the dimension variable, we consider a fully-connected deep neural network with the same size for different hidden layers to learn MNIST dataset.



As shown in Fig. \ref{fig:dnnmnist}, at initialization, the peak of RDF for a deeper hidden layer locates at a higher frequency. As the training goes, the peak of RDF of a deeper hidden layer moves towards low frequency faster. At the end of the training, the frequency of the RDF peak monotonically decreases as the hidden layer gets deeper. This indicates that the effective target function for a deeper hidden layer evolves faster towards a low frequency function, i.e., the deep frequency principle in a deep neural network.

\section{Discussion}
In this work, we empirically show a deep frequency principle that provides a promising mechanism for understanding the effect of depth in deep learning, that is, the effective target function for a deeper hidden layer biases towards  lower  frequency during the training.  Specifically, based on the well-studied frequency principle, the deep frequency principle well explains why deeper learning can be faster. We believe that the study of deep frequency principle would provide valuable insight for further theoretical understanding of deep neural networks. Next, we will discuss the relation of this work to other studies and some implications.
\subsection{Kernel methods}
Kernel methods, such as support vector machine and random feature model, are powerful at fitting non-linearly separable data. An intuitive explanation is that when data are projected into a much higher dimensional space, they are closer to be linearly separable. From the perspective of Fourier analysis, we quantify this intuition through the low-frequency ratio. After projecting data to higher dimensional space through the hidden layers, neural networks transform the high-dimensional target function into a lower frequency effective function. The deeper neural networks project data more than once into high dimensional space, which is equivalent to the combination of multiple kernel methods. In addition, neural networks not only learn the weights of kernels but also are able to learn the kernels, showing a much more capability compared with kernel methods.

\subsection{Generalization}
Frequency principle reveals a low-frequency bias of deep neural networks \cite{xu_training_2018,xu2019frequency}, which provides qualitative understandings \cite{xu2018understanding,zhang2019explicitizing} for the good generalization of neural networks in problems dominated by low frequencies, such as natural image classification tasks, and for the poor generalization in problems dominated by high frequencies, such as predicting parity function. Generalization in real world problems \cite{he2016deep} is often better as the network goes deeper. How to characterize the better generalization of deeper network is also a critical problem in deep learning. This work, validating a deep frequency principle, may provide more understanding to this problem in future work.  As the network goes deeper, the effective target function for the last hidden layer is more dominated by low frequency. This deep frequency principle phenomenon is widely observed, even in fitting high-frequency function, such as parity function in our experiments. This suggest that deeper network may have more bias towards low frequency. However, it is difficult to examine the frequency distribution of the learned function on the whole Fourier domain due to the high dimensionality of data. In addition, since the generalization increment of a deeper network is more subtle, we are exploring a more precise characterization of the frequency distribution of a high-dimensional function. 

\subsection{How deep is enough?}
The effect of depth can be intuitively understood as a pre-condition that transforms the target function to a lower frequency function. Qualitatively, it requires more layers to fit a higher frequency function. However, the effect of depth can be saturated. For example, the effective target functions for very deep layers can be very similar in the Fourier domain (dominated by very low frequency components) when the layer number is large enough, as an example shown in Fig. \ref{fig:dnnmnist}(g, h). A too deep network would cause extra waste of computation cost. A further study of the deep frequency principle may also provide a guidance for design the depth of the network structure.

\subsection{Vanishing gradient problem}
\begin{figure}[ht]
    \begin{centering}
    \subfloat[]{
        \includegraphics[scale=0.58]{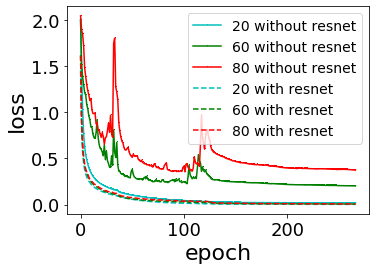}
    }
    \subfloat[]{
        \includegraphics[scale=0.58]{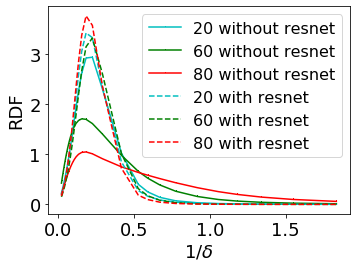} 
    }
    \par\end{centering}
    \caption{(a) Training loss vs. epoch. (b) RDF of $S^{[-3]}$  vs. $1/\delta$ at epoch 267. Legend is the number of layers of the fully-connected  network with (``with resnet'') or without (``without resnet'') residual connection  trained to fit MNIST.   }\label{fig:dnnmnistdeep}
\end{figure}

When the network is too deep, vanishing gradient problem often arises, slows down the training, and deteriorates the generalization \cite{he2016deep}. As an example, we use very deep fully connected networks, i.e., 20,  60 and 80 layers, to learn MNIST dataset. As shown in Fig. \ref{fig:dnnmnistdeep}(a)  (solid lines), deeper networks learn slower in such very deep situations. The frequency distribution of effective target function also violates deep frequency principle in this case. For example, at epoch 267 as shown in Fig. \ref{fig:dnnmnistdeep}(b) (solid lines), the RDFs of $S^{[-3]}$ of different networks show that the effective target function of a deeper hidden layer has more power on high-frequency components. Residual connection is proposed to save deep neural network from vanishing gradient problem and utilize the advantage of depth  \cite{he2016deep}, in which deep frequency principle is satisfied as shown in the previous example in Fig. \ref{fig:resnetpdf}. To verify the effectiveness of residual connection, we add residual connection to the very deep fully connected networks to learn MNIST dataset.  As shown in Fig. \ref{fig:dnnmnistdeep}(a) (dashed lines), the learning processes for different networks with residual connections are almost at the same speed. The RDFs in Fig. \ref{fig:dnnmnistdeep}(b)  (dashed lines) show that with residual connections, the depth only incurs almost a negligible effect of deep frequency principle, i.e., a saturation phenomenon. The detailed study of the relation of the depth and the difficulty of the task is left for further research.

Taken together, the deep frequency principle proposed in this work may have fruitful implication for future study of deep learning. A detailed study of deep frequency principle may require analyze different dynamical regimes of neural networks. As an example, a recent work \cite{luo2020phase} draws a phase diagram for two-layer ReLU neural networks at infinite width limit by three regimes, linear, critical and condensed regimes. Such study could inspire the study of phase diagram of deep neural networks. The linear regime is well studied \cite{jacot2018neural,zhang2019explicitizing,zhang2020type,arora2019exact}, which may be a good starting point and shed lights on the study of other regimes.
 
\section{Acknowledgements}
Z.X. is supported by National Key R\&D Program of China (2019YFA0709503), Shanghai Sailing Program, Natural Science Foundation of Shanghai (20ZR1429000), NSFC 62002221 and partially supported by HPC of School of Mathematical Sciences and Student Innovation Center at Shanghai Jiao Tong University.

\bibliographystyle{plain}
\bibliography{DLRef}

\end{document}